\documentclass[conference]{IEEEtran}
\IEEEoverridecommandlockouts
\usepackage{cite}
\usepackage{amsmath,amssymb,amsfonts}
\usepackage{algorithmic}
\usepackage{graphicx}
\usepackage{textcomp}
\usepackage{xcolor}

\usepackage{longtable,rotating}
\usepackage{threeparttable}
\usepackage{subfigure}
\usepackage{graphicx}
\usepackage{amssymb}
\usepackage{lineno}
\usepackage{multirow}
\usepackage{bm}
\usepackage[ruled,linesnumbered]{algorithm2e}

\def\BibTeX{{\rm B\kern-.05em{\sc i\kern-.025em b}\kern-.08em
    T\kern-.1667em\lower.7ex\hbox{E}\kern-.125emX}}

\begin{document}

\title{Cost-Quality Adaptive Active Learning for Chinese Clinical Named Entity Recognition}

\author{
\IEEEauthorblockN{Tingting Cai$^1$, Yangming Zhou$^{1,2,*}$ and Hong Zheng$^{1,*}$}
\IEEEauthorblockA{
$^1$School of Information Science and Engineering, East China University of Science and Technology, Shanghai 200237, China \\
$^2$Key Laboratory of Advanced Control and Optimization for Chemical Processes, Ministry of Education, China \\
$^*$Corresponding authors\\
Emails: \{ymzhou,zhenghong\}@ecust.edu.cn
}
}

\maketitle

\begin{abstract}

Clinical Named Entity Recognition (CNER) aims to automatically identity clinical terminologies in Electronic Health Records (EHRs), which is a fundamental and crucial step for clinical research. To train a high-performance model for CNER, it usually requires a large number of EHRs with high-quality labels. However, labeling EHRs, especially Chinese EHRs, is time-consuming and expensive. One effective solution to this is active learning, where a model asks labelers to annotate data which the model is uncertain of. Conventional active learning assumes a single labeler that always replies noiseless answers to queried labels. However, in real settings, multiple labelers provide diverse quality of annotation with varied costs and labelers with low overall annotation quality can still assign correct labels for some specific instances. In this paper, we propose a Cost-Quality Adaptive Active Learning (CQAAL) approach for CNER in Chinese EHRs, which maintains a balance between the annotation quality, labeling costs, and the informativeness of selected instances. Specifically, CQAAL selects cost-effective instance-labeler pairs to achieve better annotation quality with lower costs in an adaptive manner. Computational results on the CCKS-2017 Task 2 benchmark dataset demonstrate the superiority and effectiveness of the proposed CQAAL.

\end{abstract}

\begin{IEEEkeywords}
Active learning, Clinical named entity recognition, Electronic health records.
\end{IEEEkeywords}

\section{Introduction}
\label{Sec:Introduction}

Clinical Named Entity Recognition (CNER) extracts patient information from unstructured Electronic Health Records (EHRs), which is an important task for further clinical research. The main goal of CNER is to identify clinical terminologies in EHRs, such as diseases, symptoms, treatments, exams and body parts. Accurate identification of these clinical concepts can provide effective decision support for patient care and treatment~\cite{li2020towards}. Compared to English texts, CNER in Chinese texts is more difficult since Chinese EHRs are recorded without explicit word delimiters. In recent years, CNER has attracted considerable research efforts, and many methods have proposed in the literature. Most of them are deep learning methods~\cite{qiu2018fast,bhatia2019joint,zhao2019neural}.

Although many advanced models have been developed for CNER, their performance still heavily depends on the manually-annotated training data. Labeling EHRs is usually time-consuming and expensive because EHRs involve many complex clinical terminologies, and only labelers with medical background are qualified for clinical annotation. It thus becomes rather difficult to train an effective model for CNER since it requires a large number of manually-annotated clinical texts.

Active learning, which iteratively selects the most informative samples for labelers to annotate, is an effective method to reduce annotation cost~\cite{dasgupta2005analysis,li2020active,cortes2019active}. It has been widely used in many Natural Language Processing (NLP) tasks, such as text classification~\cite{orth2020early} and event recognition~\cite{hasan2020context}. In conventional active learning, there is only one labeler and the algorithm queries the labels of the selected instances from the labeler, which always returns the ground truth of queried labels~\cite{huang2017cost}. However, in many real settings, there are multiple labelers, and they usually provide diverse quality of annotation with different costs. Obviously, a labeler which offers better overall quality will require a higher cost for each query. The overall quality of labelers can be assessed according to their previous annotation  performance. Moreover, labelers may have diverse expertise for different instances. For example, in CNER tasks, some labelers may be good at labeling diseases, while some are skilled in symptoms. Therefore, we need to consider querying which of them to annotate the selected instances so as to keep a trade-off between quality and cost.

In the past few years, active learning with multiple noisy labelers has received significant attention and achieved great success in various applications. However, many works either ignored the different expertise of multiple labelers and queried the same labeler for all instances globally~\cite{zheng2010active,zhang2014active} or neglected the annotation costs of different labelers~\cite{fang2014active,yan2011active}. Recently, two methods considering the diversity of labelers on both expertise and query costs have been proposed for classification tasks~\cite{huang2017cost,gao2020cost}. Experimental results demonstrate the effectiveness of the two methods on selecting cost-effective queries. We thus follow the trend and focus on CNER task for the first time.

In this paper, we propose a Cost-Quality Adaptive Active Learning (CQAAL) method for CNER in Chinese EHRs, which selects the most cost-effective instance-labeler pairs to obtain better annotation performance with lower costs in an adaptive manner. Specifically, we first combine three sampling strategies, namely uncertainty, entropy and margin to assess the informativeness of instances. We further observe that a labeler with low quality of overall annotation can still assign accurate labels for some specific instances in real settings. Then, based on this fact, for each instance, we select a suitable labeler which offers high-quality yet cheap annotations so as to keep a balance between the annotation quality, labeling costs, and the informativeness of instances.

The main contributions of this paper can be summarized as follows:
\begin{itemize}
    \item
    We propose an active learning method for CNER, called CQAAL, which can effectively relieve the efforts of Chinese EHRs annotation. CQAAL selects instance-labeler pairs to obtain better annotation quality with lower costs and keeps a trade-off between annotation quality, labeling costs, as well as informativeness of selected instances evaluated through uncertainty, entropy and margin sampling strategies.
    \item We conduct experiments on the CCKS-2017 Task 2 benchmark dataset for CNER, extensive experimental results show that compared to baselines, our proposed CQAAL achieves highly-competitive performance, which indicates the superiority and effectiveness of CQAAL on selecting cost-effective instance-labeler pairs to maintain a balance between the annotation quality, labeling costs, and the informativeness of selected instances.
\end{itemize}

The rest of this paper is organized as follows. Section \ref{Sec:Related Work} briefly reviews the related work on CNER and active learning. Section \ref{Sec:Proposed Method} presents our proposed cost-quality adaptive active learning method for Chinese CNER, followed by experimental evaluations as Section \ref{Sec:Experiments and Analysis}. Finally, the conclusions and potential research
directions are summarized as Section \ref{Sec:Conclusion and Future Work}.

\section{Related work}
\label{Sec:Related Work}

\subsection{Clinical Named Entity Recognition}
Due to the practical significance, CNER has been studied extensively~\cite{wang2019incorporating,chen2020towards,el2019embedding}. Generally, CNER can be formulated as a clinical sequence labeling task with BIO (Begin, Inside, Outside) tags. Existing methods for CNER can be roughly divided into two categories: statistical machine learning methods and deep learning methods.

\textbf{Statistical machine learning methods}, such as hidden markov models~\cite{zhang2004enhancing,zhao2004named}, maximum entropy markov models~\cite{lin2004maximum,mccallum2000maximum}, Conditional Random Fields (CRF)~\cite{feng2011intelligent,settles2004biomedical} and Support Vector Machines (SVM)~\cite{ju2011named,makino2002tuning}, were widely used for CNER.
For example, Feng et al.~\cite{feng2011intelligent} employed CRF to recognize and classify entities in Chinese EHRs. Ju et al.~\cite{ju2011named} used SVM to NER in biomedical texts.
However, these methods heavily rely on manual feature engineering, while handcrafted features are difficult to pre-define.

Compared to the above-mentioned methods, \textbf{deep learning methods} have attracted increasing attention since they do not rely on feature engineering and learn feature representations automatically. For example, Qiu et al.~\cite{qiu2018fast} proposed a Residual Dilated Convolutional Neural Network with Conditional Random Field (RD-CNN-CRF) to capture the contextual information, and employed residual connections to utilize both semantic (i.e., high-level) and low-level features.
Wang et al.~\cite{wang2019incorporating} incorporated dictionaries into Bi-directional Long Short-Term Memory (BiLSTM) to combine data-driven deep learning approaches and knowledge-driven dictionary approaches. Li et al.~\cite{li2020chinese} pre-trained BERT model on the unlabeled Chinese clinical records, which can leverage the unlabeled domain-specific knowledge. Li et al.~\cite{li2020towards} proposed a dynamic embedding method based on dynamic attention which combines features of both characters and words in embedding layers for CNER. Chen et al~\cite{chen2020towards} proposed a novel framework stacking the Bayesian network ensembles on top of the entity-aware CNN to automatically identify diagnosis.

To train an effective CNER model, large numbers of manually-annotated clinical texts are usually required. However, annotating EHRs is often time-consuming and expensive. Active learning can solve this problem to a certain extent, where a model asks labelers to annotate data that the model is uncertain of iteratively.

\begin{figure*}[!ht]
\begin{center}
\includegraphics[width=0.8\textwidth]{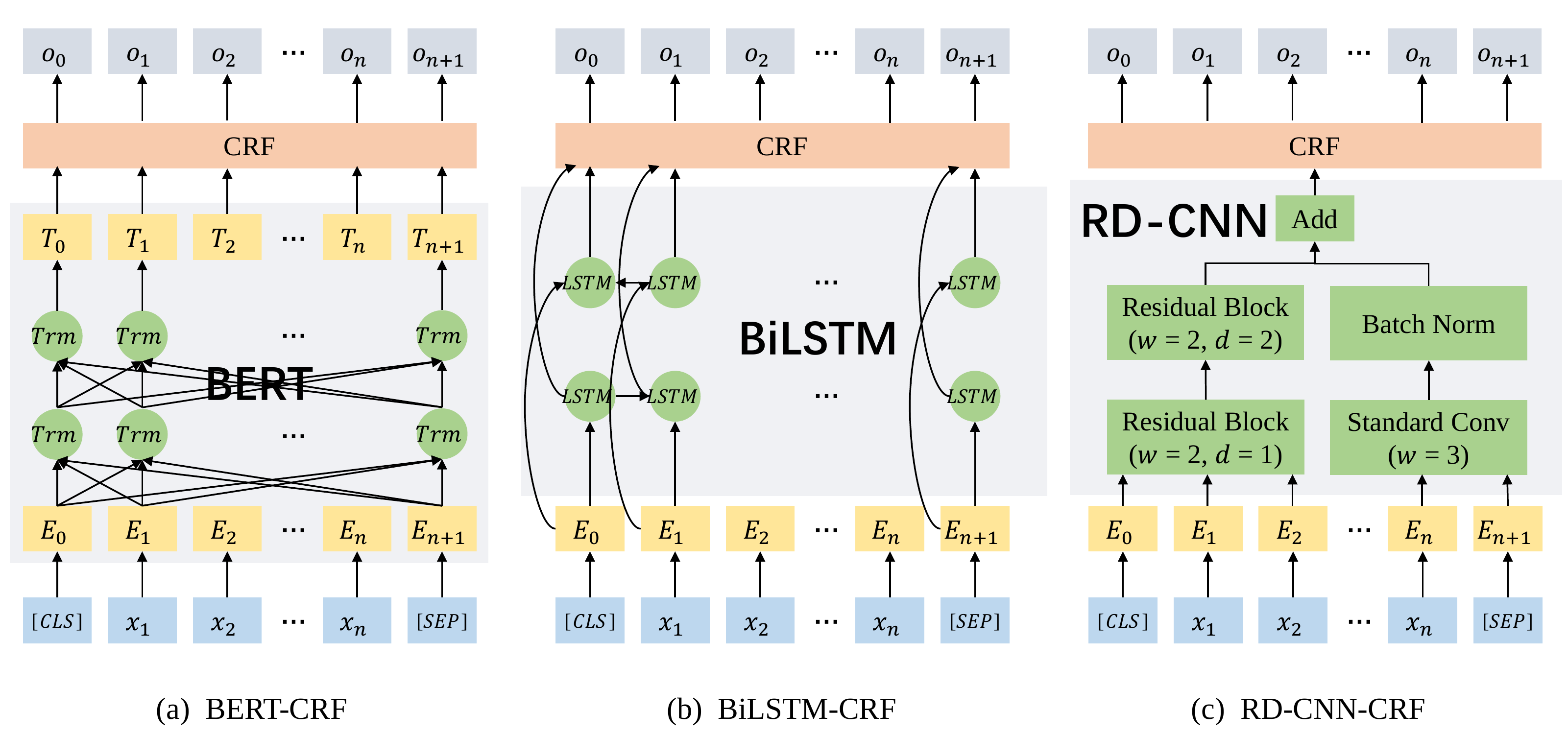}
\caption{The detailed architectures of labelers, where $x_i$ means the $i^{th}$ character, and $o_i$ means the output of $x_i$.}
\label{fig:Model}
\end{center}
\end{figure*}

\subsection{Active Learning}
Active learning aims to reduce the annotation cost by
actively selecting the most informative instances to be queried~\cite{yan2020active}. Over the past few years, active learning has been well studied and shown to be effective for various tasks~\cite{cohn1994improving,dasgupta2005analysis,li2020active}. Researches of active learning can be categorized into two groups: active learning with a single labeler and active learning with multiple labelers.

\textbf{Active learning with a single labeler} assumes that there is only one labeler which always assigns accurate labels for the selected instances. This method devotes to selecting valuable samples which can improve model performance to the largest extent, such as uncertainty sampling~\cite{culotta2005reducing}, query-by-committee~\cite{seung1992query}, and fisher information~\cite{settles2008analysis}. However, in real-world settings, we usually have multiple labelers and it is almost impossible for all the labelers to annotate perfectly all the time. They tend to provide diverse annotation quality, thus charging different costs.

In recent years, \textbf{active learning with multiple labelers} has received more and more research interests.
For instance, Zheng et al.~\cite{zheng2010active} presented a new algorithm IEAdjCost for active learning from multiple labelers with unknown, varied accuracies and known, varied costs. However, they assume that all labelers have the same expertise and query one labeler for all instances globally.
In addition, some researches which consider the diverse expertise of different labelers have been conducted.
For example, Fang et al.~\cite{fang2014active} proposed to transfer knowledge from auxiliary domains to estimate labelers expertise in active learning for crowdsourcing.
Ambati et al.~\cite{ambati2010active} actively selected both instances and labelers for machine translation.
However, they ignored the varied annotation costs of different labelers.
Currently, there are a few attempts to exploit the expertise and costs of labelers for better annotation with lower costs. Huang et al.~\cite{huang2017cost} proposed a novel active selection criterion to evaluate the cost-effectiveness of instance-labeler pairs, which brings us inspiration. However, they focused on the classification tasks and evaluated the informativeness of instances based solely on uncertainty. Moreover, they conducted experiments with real labelers.

Compared to Huang et al.~\cite{huang2017cost}, we investigate the effectiveness of cost-quality adaptive active learning method in CNER for the first time. To select the instances which are beneficial to model performance improvement, we evaluate the informativeness of instances from three perspectives, namely uncertainty, entropy and margin. Furthermore, in our experiments, we employ neural network models with different architectures as diverse labelers to reduce the cost of experiments.

\section{Proposed Method}
\label{Sec:Proposed Method}

Given an initial labeled dataset $\mathcal{L}=\left\{\mathbf{x}_{j}, y_{j}\right\}_{j=1}^{n}$, a training set $\mathcal{D}=\left\{\mathbf{x}_{j}, y_{j}\right\}_{j=n+1}^{t}$, a large pool of unlabeled data $\mathcal{U}=\left\{\mathbf{x}_{j}\right\}_{j=t+1}^{N}$, and a set of neural network models $\mathcal{M}=\left\{m_{i}\right\}_{i=1}^{z}$, we first use training set $\mathcal{D}$ to train models in $\mathcal{M}$. In our experiments, we utilize models with different architectures to simulate labelers in real settings. There are three labelers (i.e., $\mathcal{M}=\left\{m_{1}, m_{2}, m_{3}\right\}$), namely BERT-CRF, BiLSTM-CRF, and RD-CNN-CRF. Fig.~\ref{fig:Model} shows the detailed architectures of our labelers. We use CRF as the decoder to make positional tagging decisions over individual characters for all labelers since CRF considers the dependencies of adjacent tags. Moreover, given a sequence, we add a special token $[CLS]$ in front of the sequence and a separator token $[SEP]$ at the end of the sequence. Then, we employ BERT tokenizer to make word segmentation for all labelers. In addition to the commonalities, they also have some respective characteristics:

\textbf{Labeler1: BERT-CRF}~\cite{devlin2018bert}. BERT is a pre-trained model, which can learn deep bidirectional representations by jointly conditioning on both left and right contexts in all layers. We fine-tune on the BERT-Base\footnote{https://github.com/google-research/bert} in Chinese, which includes 12 layers, 768 hiddens, 12 heads and 110M parameters.

\textbf{Labeler2: BiLSTM-CRF}~\cite{hochreiter1997long}. BiLSTM introduced gate mechanism to solve the gradient disappearing problem of traditional Recurrent Neural Network (RNN). It can model the whole sequence and also capture the long-term dependencies within the sequence~\cite{qin2020feature}.

\textbf{Labeler3: RD-CNN-CRF}~\cite{qiu2018fast}. CNN is able to extract some local and position-invariant features, which are important for CNER~\cite{yin2017comparative}. Residual connections are utilized to capture both semantic and contextual features to generate better representations.

In real settings, each labeler provides different quality of annotation, and the labeler with higher overall performance accordingly requires higher costs. Moreover, each labeler has its own expertise, and one achieves low overall results may assign accurate labels for some specific instances. Therefore, we select a suitable labeler for each instance, i.e., we select instance-labeler pairs with the following properties:
1) The instance is informative, valuable, and beneficial to entity recognition model improvement.  2) The cost of the selected labeler is low. 3) The selected labeler can annotate the instance as accurately as possible, i.e., the quality of the annotation is great.

At each iteration of active learning, we select $k$ instance-labeler pairs $\left\{(\mathbf{x}^{*}_{j}, m^{*}_{i})_l \mid \mathbf{x}^{*}_{j} \in \mathcal{U}, m^{*}_{i} \in \mathcal{M}\right\}_{l=1}^{k}$, and query the labels of $\left\{(\mathbf{x}^{*}_{j})_l\right\}_{l=1}^{k}$ from the labelers $\left\{(m^{*}_{i})_l\right\}_{l=1}^{k}$.
Below, we first introduce how we measure the informativeness of instances, the quality of annotation, and the cost of diverse labelers. Then, we define the evaluation of cost-quality by combining them together. Finally, we demonstrate the algorithm of our proposed CQAAL.

\subsection{The Informativeness of Instances}
The core idea of active learning is to select the most informative instances for the current model. Sampling strategies play an important role, and have been well studied over the past few years~\cite{lewis1994heterogeneous,marcheggiani2014experimental}. In sequence labeling tasks, uncertainty, entropy and margin sampling strategies are widely used due to their simplicity yet competitive performance~\cite{culotta2005reducing,settles2008analysis,balcan2007margin}.

\textbf{Uncertainty}~\cite{culotta2005reducing}. If the current entity recognition model is uncertain about its prediction on an instance, then the instance may be more useful for improving the model performance since it includes more information which the model has not learnt yet. The uncertainty of an instance can be calculated as follows.
\begin{equation}
{Uncertainty}(\mathbf{x}) = 1 - \max _{y \in \mathcal{Y}} p\left(y \mid \mathbf{x}\right)
\end{equation}
where $p\left(y \mid \mathbf{x}\right)$ means the probability that $\mathbf{x}$ belongs to the class $y$, and $\mathcal{Y}$ represents the set of all classes.

\textbf{Entropy}~\cite{settles2008analysis}. Entropy can evaluate the ambiguity about
the label of an instance. If the distribution of marginal probability uniforms close, the entropy of the instance is large.
\begin{equation}
{Entropy}(\mathbf{x})= - \sum_{y \in \mathcal{Y}} p\left(y \mid \mathbf{x}\right) \cdot \log p\left(y \mid \mathbf{x}\right)
\end{equation}

\textbf{Margin}~\cite{balcan2007margin}. Margin considers the two most likely assignments, which is calculated by subtracting the highest probability by the second one. If the margin of an instance is small, the current entity recognition model is unsure about the decision.
\begin{equation}
{Margin}(\mathbf{x}) = \max _{y \in \mathcal{Y}} p\left(y \mid \mathbf{x}\right)-\max _{y \in \mathcal{Y}}^{\prime} p\left(y \mid \mathbf{x}\right)
\end{equation}
where $\max _{y \in \mathcal{Y}}^{\prime} p\left(y \mid \mathbf{x}\right)$ denotes the second maximum probability.

We sum up the three sampling strategies to evaluate the informativeness of an instance from different perspectives:
\begin{equation}
{S}(\mathbf{x}) = \\
{Uncertainty}(\mathbf{x}) + \frac{Entropy(\mathbf{x})}{\log{N}} + (1-{Margin}(\mathbf{x}))
\end{equation}
Note that we normalize the entropy by $\log{N}$, where $N$ denotes the number of classes. Furthermore, we substrate ${Margin}(\mathbf{x})$ from 1 to use the $max$ operation with the margin of an instance.

Finally, we average ${S}(\mathbf{x})$ to measure the informativeness of an instance:
\begin{equation}
{Informativeness}(\mathbf{x}) = \frac{{S}(\mathbf{x})}{3}
\end{equation}

\subsection{The Quality of Annotation}
As discussed before, the quality of a labeler on a specific instance cannot be evaluated by its overall annotation quality. When estimating the quality of annotation, we follow the work of~\cite{huang2017cost}. We assume that a labeler will make similar annotation for similar instances, thus achieving similar performance. Therefore, given an instance $\mathbf{x}$, if a labeler assigns accurate labels
to the neighbors of $\mathbf{x}$ in the labeled set, then its annotation on $\mathbf{x}$ can be considered reliable. Formally, the annotation quality of labeler ${m}_{i}$ on the instance $\mathbf{x}_{j}$ can be estimated as follows.
\begin{equation}
{Quality}_{i}\left(\mathbf{x}_{j}\right)=\frac{1}{a} \sum_{\mathbf{x}_{h} \in N\left(\mathbf{x}_{j}, a\right)} S\left(\mathbf{x}_{j}, \mathbf{x}_{h}\right) G\left(y_{h}, y_{ih}\right)
\end{equation}
where $N\left(\mathbf{x}_{j}, a\right)$ represents a set of $a$ nearest neighbors of $\mathbf{x}_{j}$ in the labeled set. $S\left(\mathbf{x}_{j}, \mathbf{x}_{h}\right)$ measures the similarity between $\mathbf{x}_{j}$ and $\mathbf{x}_{h}$. Here, we select neighbors through cosine similarity. $G\left(y_{h}, y_{ih}\right)$ is an indicator function, which can be calculated as follows.
\begin{equation}
G\left(y_{h}, y_{ih}\right) = \left\{\begin{array}{ll}
1 & \text { if }  y_{h} == y_{ih}\\
0 & \text { else }
\end{array}\right.
\end{equation}
where $y_{h}$ denotes the true label of $\mathbf{x}_{h}$, and $y_{ih}$ means the prediction on instance $\mathbf{x}_{h}$ from labeler ${m}_{i}$. It is obvious that,  among the $a$ neighbors, the instances which are more similar to $\mathbf{x}_{j}$ have more contribution to the evaluation of ${Quality}_{i}\left(\mathbf{x}_{j}\right)$.

\subsection{The Cost of Diverse Labelers}
Obviously, if a labeler provides great overall quality of annotation, it will require a high cost for each query. We evaluate the cost of diverse labelers by their performance on the initial labeled data $\mathcal{L}$. The cost of labeler ${m}_{i}$ for answering each query can be formalized as follows.
\begin{equation}
{Cost}\left({m}_{i}\right)=H\left(F_1(m_{i}, \mathcal{L})\right)
\end{equation}
where $F_1(m_{i}, \mathcal{L})$ represents the $F_1$-score of labeler ${m}_{i}$ on $\mathcal{L}$. $H(\cdot)$ is an increasing function which indicates the relation between the cost and $F_1$-score.

\subsection{The Evaluation of Cost-Quality}
To select suitable instance-labeler pairs, we want the ${Informativeness}(\mathbf{x}_{j})$ to be large, the ${Quality}_{i}\left(\mathbf{x}_{j}\right)$ to be high, and the ${Cost}\left({m}_{i}\right)$ to be low. Based on the previous formulas, an evaluation function of the cost-quality about the $\left\{\mathbf{x}_{j}, {m}_{i}\right\}$ can be defined as follows.
\begin{equation}
E\left(\mathbf{x}_{j}, m_{i}\right)=\frac{{Informativeness}(\mathbf{x}_{j}) \cdot {Quality}_{i}\left(\mathbf{x}_{j}\right)}{{Cost}\left({m}_{i}\right)}
\end{equation}

We can see that only the instance-labeler pairs which meet the above three conditions can achieve high evaluation scores.

The final selection of instance-labeler pair is formulated as:
\begin{equation}
\left(\mathbf{x}^{*}, m^{*}\right)=\underset{\mathbf{x}_{j} \in \mathcal{U}, m_{i} \in \mathcal{M}}{\arg \max } E\left(\mathbf{x}_{j}, m_{i}\right)
\end{equation}

\subsection{Cost-Quality Adaptive Active Learning for CNER}

The pseudo code of the proposed CQAAL is demonstrated in Algorithm~\ref{alg:1}.
First, we use $\mathcal{D}$ to train models in $\mathcal{M}$, and define the costs for all labelers (i.e., neural network models) according to their performance on initial labeled data $\mathcal{L}$.
At each iteration of active learning, the algorithm selects $k$ instance-labeler pairs $\left\{(\mathbf{x}^{*}_{j}, m^{*}_{i})_l \mid \mathbf{x}^{*}_{j} \in \mathcal{U}, m^{*}_{i} \in \mathcal{M}\right\}_{l=1}^{k}$, where the instances are helpful to improve the performance of entity recognition model, and the labelers can assign correct labels for the selected instances with low costs. Then, the algorithm queries the labels of $\left\{(\mathbf{x}^{*}_{j})_l\right\}_{l=1}^{k}$ from the corresponding labelers $\left\{(m^{*}_{i})_l\right\}_{l=1}^{k}$. Later, $\left\{(\mathbf{x}^{*}_{j})_l\right\}_{l=1}^{k}$ are removed from the unlabeled dataset $\mathcal{U}$, and are incorporated into labeled datatset $\mathcal{L}$ along with queried labels $\left\{(y^{*}_{j})_l\right\}_{l=1}^{k}$ from $\left\{(m^{*}_{i})_l\right\}_{l=1}^{k}$. Afterwards, the entity recognition model is retrained with the new labeled data $\mathcal{L}$, and we evaluate the model on the test set. In our work, we employ Bi-directional Gated Recurrent Unit with Conditional Random Field (BiGRU-CRF) as the entity recognition model. GRU~\cite{chung2015gated} is a popular variant of LSTM, which includes reset gate and update gate. Compared to LSTM, GRU can support longer sequences and is more simple~\cite{lei2018effective}.
Finally, the above steps of active learning iterate until the desired $F_{1}$-score is reached or the number of iterations has reached a predefined threshold.

\begin{algorithm}[!ht]
\caption{The Pseudo-code of CQAAL Algorithm}
\label{alg:1}
\KwIn{\\
\qquad$\mathcal{L}$: Initial labeled data,\\
\qquad$\mathcal{D}$: Training set to train labelers,\\
\qquad$\mathcal{U}$: The pool of unlabeled data,\\
\qquad$\mathcal{M}$: A set of neural network models,\\
\qquad$M$: The number of iterations.
}
\Begin{
\textbf{Initialize:}\\\qquad
    Use $\mathcal{D}$ to train models in $\mathcal{M}$\\
    \qquad
    Calculate the costs for all labelers (i.e., neural network models) based on Equation (8)\\
\For{$index=1$ to $M$}{
    \For{ each $\mathbf{x}_{j} \in \mathcal{U}$ and labeler $m_{i} \in \mathcal{M}$ }{
    Compute $Informativeness$ of $\mathbf{x}_{j}$ according to Equation (5)\\
    Calculate the $Quality$ of $m_{i}$ on $\mathbf{x}_{j}$ based on Equation (6)\\
    Calculate the cost-quality of $\left(\mathbf{x}^{*}, m^{*}\right)$ as Equation (9)\\
    }
    Select $k$ instance-labeler pairs $\left\{(\mathbf{x}^{*}_{j}, m^{*}_{i})_l \mid \mathbf{x}^{*}_{j} \in \mathcal{U}, m^{*}_{i} \in \mathcal{M}\right\}_{l=1}^{k}$ based on Equation (10)\\
    Query the labels of $\left\{(\mathbf{x}^{*}_{j})_l\right\}_{l=1}^{k}$ from the labelers $\left\{(m^{*}_{i})_l\right\}_{l=1}^{k}$, denoted by $\left\{(y^{*}_{j})_l\right\}_{l=1}^{k}$\\
    Form a new labeled data $\mathcal{L}\leftarrow \mathcal{L} \bigcup \left\{(\mathbf{x}^{*}_{j}, {y}^{*}_{j} )_l\right\}_{l=1}^{k}$\\
    Form a new unlabeled data $\mathcal{U} \leftarrow \mathcal{U} \backslash \left\{(\mathbf{x}^{*}_{j})_l\right\}_{l=1}^{k}$\\
    Train an entity recognition model on $\mathcal{L}$\\
    Evaluate the model on the test set\\
}
}
\end{algorithm}

\section{Experiments and Analysis}
\label{Sec:Experiments and Analysis}
We compare our proposed CQAAL with the following baseline methods:
\begin{itemize}
\item \textbf{CEAL:} CEAL~\cite{huang2017cost} evaluates the informativeness of instances solely based on uncertainty.
\item \textbf{AQ:} selecting instances actively and always querying the labeler with the highest overall quality.
\item \textbf{RQ:} selecting instances randomly and always querying the labeler with the highest overall quality.
\item \textbf{AC:} selecting instances actively and always querying the labeler with the lowest cost.
\item \textbf{RC:} selecting instances randomly and always querying the labeler with the lowest cost.
\end{itemize}

The main goal of the evaluation is to compare the effectiveness of different algorithms on choosing cost-effective instance-labeler pairs. BiGRU-CRF is used as the entity recognition model in our experiments.

\subsection{Dataset}
\label{SubSec:Dataset}

We use the CCKS-2017 Task 2 benchmark dataset\footnote{https://www.biendata.com/competition/CCKS2017\_2} to conduct our experiments. The dataset contains 1,596 annotated instances with five types of clinical named entities, including diseases, symptoms, exams, treatments, and body parts. The annotated instances are already
partitioned into 1,198 training instances and 398 test instances. We divide the 1,198 training instances into initial labeled data, unlabeled data, and training data which is used to train labelers (i.e., neural network models). The statistics of different types of clinical named entities are listed in Table~\ref{tab:statistics}.

\begin{table}[!ht]
\begin{center}
\caption{statistics of different types of clinical named entities}
\label{tab:statistics}
\begin{tabular}{l|lllr}
\hline
Type & Initial labeled & Training & Unlabeled & Test\\ \hline
Diseases & 169  & 172 & 381 & 516 \\ \hline
Symptoms & 1,566  & 1,664 & 4,601 & 2,257 \\ \hline
Exams & 1,851  & 2,048 & 5,647 & 3,013 \\ \hline
Treatments & 204  & 241 & 603 & 451 \\ \hline
Body parts & 2,121  & 2,286 & 6,312 & 2,875 \\ \hline
\end{tabular}
\end{center}
\end{table}

\subsection{Parameter Settings}
\label{SubSec:Parameter Settings}

The configuration of parameters may influence the performance of a neural network model. The parameter settings of the
proposed algorithm are shown in Table~\ref{tab:parameters}.
For BERT-CRF, we use the default parameters of BERT-Base in Chinese. Each model runs on a single NVIDIA GeForce GTX 1080 Ti GPU. We use the Adaptive Moment Estimation (Adam)~\cite{kingma2014adam}  optimization algorithm in training.

\begin{table}[!ht]
\begin{center}
\caption{Parameter Settings}
\label{tab:parameters}
\begin{tabular}{l|lr}
\hline
Model & Parameter & Value \\ \hline
\multirow{3}{*}{All} & Maximum sequence length & $len$ = 128 \\ \cline{2-3}
 & Size of word embedding & $d_w$ = 128 \\ \cline{2-3}
 & Learning rate & $lr$ = 0.001\\ \hline
\multirow{6}{*}{RD-CNN-CRF} & Number of residual block & $n_r$ = 2 \\ \cline{2-3}
 & Number of filters per residual block & $f_r$ = 256 \\ \cline{2-3}
 & Window size of dilated convolution & $w_d$ = 2 \\ \cline{2-3}
 & Dilation factor of the $i_th$ residual block & $d = 3^{i-1}$ \\ \cline{2-3}
 & Number of filters for standard convolution & $f_s$ = 256 \\ \cline{2-3}
 & Window size of standard convolution & $w_s$ = 3 \\ \hline
BiLSTM-CRF & Number of LSTM hidden units & $n_l$ = 256 \\  \hline
BiGRU-CRF & Number of GRU hidden units & $n_g$ = 256 \\ \hline
\end{tabular}
\end{center}
\end{table}

\begin{table*}[!htb]
\begin{center}
\caption{performance on the initial labeled data $\mathcal{L}$ given by each labeler.}
\label{tab:PerformanceOfEachLabeler}
\begin{tabular}{l|ccc|ccc|ccc}
\hline
\multirow{2}{*}{} & \multicolumn{3}{c|}{BERT-CRF} & \multicolumn{3}{c|}{BiLSTM-CRF} & \multicolumn{3}{c}{RD-CNN-CRF} \\ \cline{2-10}
 & \multicolumn{1}{c}{Precision} & \multicolumn{1}{c}{Recall} & \multicolumn{1}{c|}{$F_1$-score} & \multicolumn{1}{c}{Precision} & \multicolumn{1}{c}{Recall} & \multicolumn{1}{c|}{$F_1$-score} & \multicolumn{1}{c}{Precision} & \multicolumn{1}{c}{Recall} & \multicolumn{1}{c}{$F_1$-score} \\ \hline
Diseases & \textbf{73.91} & \textbf{77.98} & \textbf{75.89} & 59.82 & 61.47 & 60.63 & 55.65 & 63.30 & 59.23 \\ \hline
Symptoms & 87.31 & \textbf{93.86} & \textbf{90.47} & \textbf{87.35} & 91.41 & 89.33 & 82.78 & 89.22 & 85.88 \\ \hline
Exams & 60.36 & \textbf{77.65} & 67.92 & \textbf{71.35} & 75.64 & \textbf{73.44} & 51.95 & 68.77 & 59.19 \\ \hline
Treatments & \textbf{45.11} & \textbf{60.61} & \textbf{51.72} & 26.35 & 39.39 & 31.58 & 23.84 & 41.41 & 30.26 \\ \hline
Body parts & \textbf{70.34} & \textbf{77.13} & \textbf{73.58} & 61.41 & 67.67 & 64.39 & 55.38 & 65.22 & 59.90 \\ \hline
Overall & \textbf{73.22} & \textbf{83.12} & \textbf{77.86} & 70.61 & 76.86 & 73.60 & 62.14 & 74.16 & 67.62 \\ \hline
\end{tabular}
\end{center}
\end{table*}

In active learning, we fix the number of iterations
at 10 since each algorithm does not improve obviously after 10 iterations. At each iteration, we select 70 instance-labeler pairs from unlabeled data $\mathcal{U}$ and neural network model sets $\mathcal{M}$ for entity recognition model BiGRU-CRF to learn. Furthermore, we set $a$ in Equation (6) as 50, i.e., given an instance $\mathbf{x}$, if a labeler can assign correct labels to 50 instances nearest to $\mathbf{x}$ from the labeled set, then its annotation on $\mathbf{x}$ can be considered reliable.

\subsection{Annotation Simulation}
\label{SubSec:Annotation Simulation}
In this experiment, we employ three models to simulate three different labelers, namely BERT-CRF, BiLSTM-CRF, and RD-CNN-CRF (see Section~\ref{Sec:Proposed Method}). Table~\ref{tab:PerformanceOfEachLabeler} presents the performance on the initial labeled data $\mathcal{L}$ given by each labeler. We observe that BERT-CRF provides the highest overall quality with the $F_1$-score of 77.86\%, while BiLSTM-CRF and RD-CNN-CRF achieve 73.60\% and 67.62\%, respectively. Moreover, we find that BiLSTM-CRF has expertise in labeling exams with the highest $F_1$-score of 73.44\% on exams, which indicates that labelers with relatively low overall quality can still achieve great performance for some specific entities. Huang et al.~\cite{huang2017cost} used the increasing function $H(m)=m$ in Equation (8) to define the cost, i.e., they set the costs as 1, 2, and 3 with overall performance from low to high. However, we find that the algorithm always selects the labeler with the least cost in such a setting. Therefore, in increasing order of annotation performance on the initial labeled data $\mathcal{L}$, we set the costs of each query from the labelers as 1, 1.2, and 1.5, respectively.

\subsection{Comparisons with Baseline Methods}
\label{SubSec:Comparisons with Baseline Methods}

\begin{figure*}
\begin{center}
\includegraphics[width=1\textwidth]{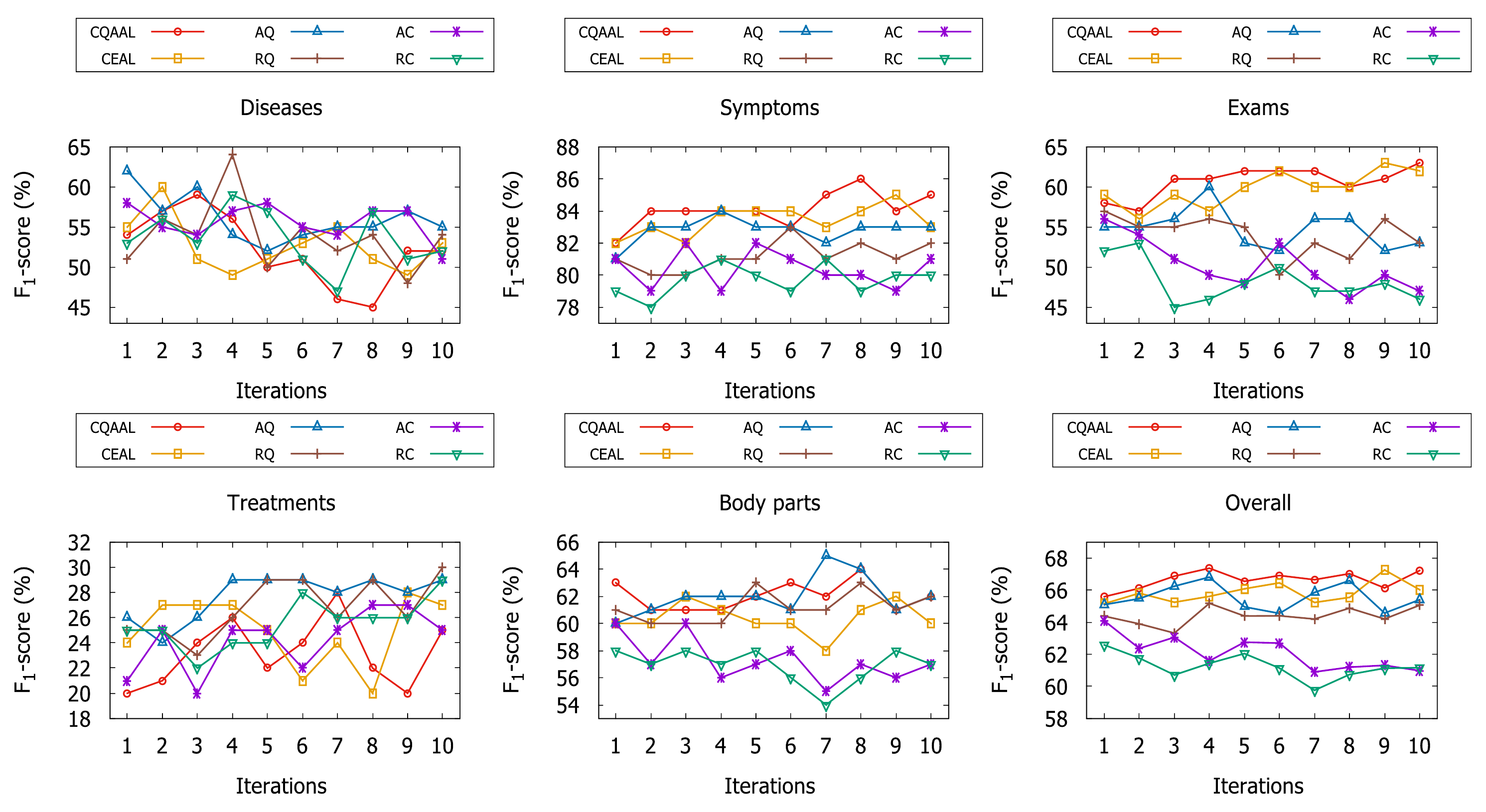}
\caption{$F_1$-score curves for active learning on specific and overall entities.}
\label{Fig:Comparisons between CQAAL and Baseline Methods}
\end{center}
\end{figure*}

We conduct this experiement as follows.
1) We train BiGRU-CRF on the initial labeled data $\mathcal{L}$, and the results are listed in Table~\ref{tab:PerformacOfBiGRUCRF}. We can see that BiGRU-CRF obtains the $F_1$-score of 62.68\% when trained on the initial labeled data $\mathcal{L}$. 2) We select instance-labeler pairs based on different algorithms and update the labeled and unlabeled data. 3) We retrain BiGRU-CRF on the new labeled data and evaluate on the test set. Step 2) and 3) iterate for 10 times. Fig.~\ref{Fig:Comparisons between CQAAL and Baseline Methods} demonstrates the comparisons between different algorithms on specific and overall entities. We find that each algorithm has different performance on different entities. For example, AQ achieves better results on treatments and body parts, while CQAAL obtains higher $F_1$-scores on symptoms and exams. Moreover, we observe that CQAAL performs best on overall entities and outperforms CEAL at most iterations since CQAAL measures the informativeness of instances on three sampling strategies (i.e., uncertainty, entropy and margin), while CEAL solely depends on uncertainty. Thus, CQAAL is able to select more valuable and informative instances than CEAL. Furthermore, it is indicated that AQ and AC achieve better performance than RQ and RC, respectively, which proves the effectiveness of active learning.

\begin{table}[!ht]
\begin{center}
\caption{performance of BiGRU-CRF trained on initial labeled data $\mathcal{L}$}
\label{tab:PerformacOfBiGRUCRF}
\begin{tabular}{l|ccc}
\hline
  & Precision & Recall & $F_1$-score \\ \hline
Diseases & 50.57  & 57.83 & 53.96 \\ \hline
Symptoms & 78.56  & 81.96 & 80.23  \\ \hline
Exams & 44.17  & 61.49 & 51.41  \\ \hline
Treatments & 20.26  & 25.13 & 22.43  \\ \hline
Body parts & 54.31  & 66.14 & 59.64  \\ \hline
Overall & 58.09  & 68.05 & 62.68 \\ \hline
\end{tabular}
\end{center}
\end{table}

We also investigate the cost-effectiveness of different algorithms. Table~\ref{tab:AverageQualityandTotalCosts} shows the average quality and costs of different algorithms in 10 iterations. CQAAL achieves the best precision of 63.50\% and $F_1$-score of 66.65\% with the lowest costs of 84.28. Compared to CEAL, the result of CQAAL is improved by 0.80\% in $F_1$-score with similar costs, which indicates the ability of CQAAL to pick up informative instances beneficial to model performance improvement. We also find that AQ obtains the highest recall of 72.41\%. However, with the highest costs of 105, AQ does not achieve the highest $F_1$-score because AQ always queries the labeler with the highest overall quality (i.e., BERT-CRF) for each instance, while BERT-CRF may make wrong annotations for some specific instances.

\begin{table}[!ht]
\begin{center}
\caption{The Average Quality and Costs of Different Algorithms in 10 Iterations}
\label{tab:AverageQualityandTotalCosts}
\begin{tabular}{l|cccr}
\hline
Algorithms  & Precision & Recall & $F_1$-score & Costs\\ \hline
CQAAL & \textbf{63.50}  & 70.19 & \textbf{66.65} & \textbf{84.28}\\ \hline
CEAL & 62.02  & 70.18 & 65.85 & 84.29 \\ \hline
AQ & 59.96  & \textbf{72.41} & 65.56 & 105.00 \\ \hline
RQ & 58.91  & 71.02 & 64.39 & 105.00 \\ \hline
AC & 56.84  & 68.49 & 62.09 & 70.00 \\ \hline
RC & 55.83  & 67.99 & 61.25 & 70.00 \\ \hline
\end{tabular}
\end{center}
\end{table}
This result proves that the labeler with highest overall quality may not be able to make correct annotations for each instance and labelers with low overall annotation quality can
still assign accurate labels for some specific instances.
Furthermore, we can see that AC and RC perform worst with the $F_1$-score of 62.09\% and 61.25\% respectively since they query all instances from the labeler with the least cost (i.e., RD-CNN-CRF). The resulting noisy annotations can yield poor model performance.

\section{Conclusion and Future Work}
\label{Sec:Conclusion and Future Work}

In this paper, we propose an active learning method with multiple labelers for CNER in Chinese EHRs, namely CQAAL. CQAAL selects instance-labeler pairs, where the selected instances are informative, the costs of selected labelers are low, and the quality of annotation is high. To evaluate the informativeness of the selected instances, we take uncertainty, entropy, and margin sampling strategies into consideration. Based on the CCKS-2017 Task 2 benchmark dataset, we experimentally evaluate our proposed CQAAL. Experimental results show that compared to baseline methods, CQAAL achieves competitive performance.
As future work, we plan to design a more complex function to evaluate the cost-effectiveness of instance-labeler pairs. Furthermore, we want to explore the application of our proposed method on other NLP tasks, such as relation extraction.

\section*{Acknowledgment}

This work was supported by National Natural Science Foundation of China (No.61772201) and National Key R\&D Program of China for ``Precision Medical Research" (No.2018YFC0910550).

\bibliographystyle{IEEEtran}
\bibliography{Bibfiles}
\end{document}